\documentclass{SCIS2024}
\usepackage{multirow}

\begin{document}
%\oa
%%%%%%%%%%%%%%%%%%%%%%%%%%%%%%%%%%%%%%%%%%%%%%%%%%%%%%%
%%% Authors do not modify the information below
%%% ×÷Õß²»ÐèÒªÐÞ¸Ä´Ë´¦ÐÅÏ¢
\ArticleType{RESEARCH PAPER}
%\SpecialTopic{}
%\luntan
\Year{2024}
\Month{}
\Vol{}
\No{}
\DOI{}
\ArtNo{}
\ReceiveDate{}
\ReviseDate{}
\AcceptDate{}
\OnlineDate{}
%%%%%%%%%%%%%%%%%%%%%%%%%%%%%%%%%%%%%%%%%%%%%%%%%%%%%%%

\title{DocPedia: unleashing the power of large multimodal model in the frequency domain for versatile document understanding}{Title keyword 5 for citation Title for citation Title for citation}

\author[1,2,$^\dagger$]{Hao~FENG}{}
\author[2,$^\dagger$]{Qi LIU}{}
\author[2,$^\dagger$,*]{Hao LIU}{haoliu.0128@bytedance.com, zhwg@ustc.edu.cn}
\author[2,$^\dagger$]{Jingqun TANG}{}
\author[1,2,*]{\\ Wengang ZHOU}{}
\author[1,2]{Houqiang LI}{}
\author[2]{Can HUANG}{}

\AuthorMark{Hao FENG}

\contributions{The first four authors contribute equally to this work. Hao Liu leads this project.}

\address[1]{University of Science and Technology of China}
\address[2]{ByteDance Inc.}

\abstract{In this work, we present DocPedia,
a novel large multimodal model (LMM) for versatile OCR-free document understanding, capable of parsing images up to 2,560$\times$2,560 resolution. Unlike existing work either struggle with high-resolution documents or give up the large language model thus vision or language ability constrained, our DocPedia directly processes visual input in the frequency domain rather than the pixel space. The unique characteristic 
enables DocPedia to capture a greater amount of visual and textual information using a limited number of visual tokens. To consistently enhance both the perception and comprehension abilities of our DocPedia, we develop a dual-stage training strategy and enrich instructions/annotations of all training tasks covering multiple document types. Extensive quantitative and qualitative experiments are conducted on various publicly available benchmarks and the results confirm the mutual benefits of jointly learning perception and comprehension tasks. The results provide further evidence of the effectiveness and superior performance of our DocPedia over other methods.}


\maketitle

\section{Introduction}
Document understanding~\cite{srihari1986document} is a critical and challenging task that sits at the intersection of the computer vision and natural language processing fields.
It involves the \textit{perception} and \textit{comprehension} in terms of visual and textual content embedded within document images.
The difficulty of this task stems from the diverse and complex formats of high-resolution documents, where the sparse or dense texts are intertwined with various graphics and tables. 
The accurate parsing of documents not only propels the digitization of archival materials but also facilitates the document automation in current data-rich world, such as information
extraction~\cite{hwang2019post,kim2022ocr,luo2023geolayoutlm} and visual question answering (VQA)~\cite{ye2023mplug,feng2023unidoc,ye2023ureader,lv2023kosmos}.

\begin{figure}[t]
	\centering
	\includegraphics[width=0.9\columnwidth]{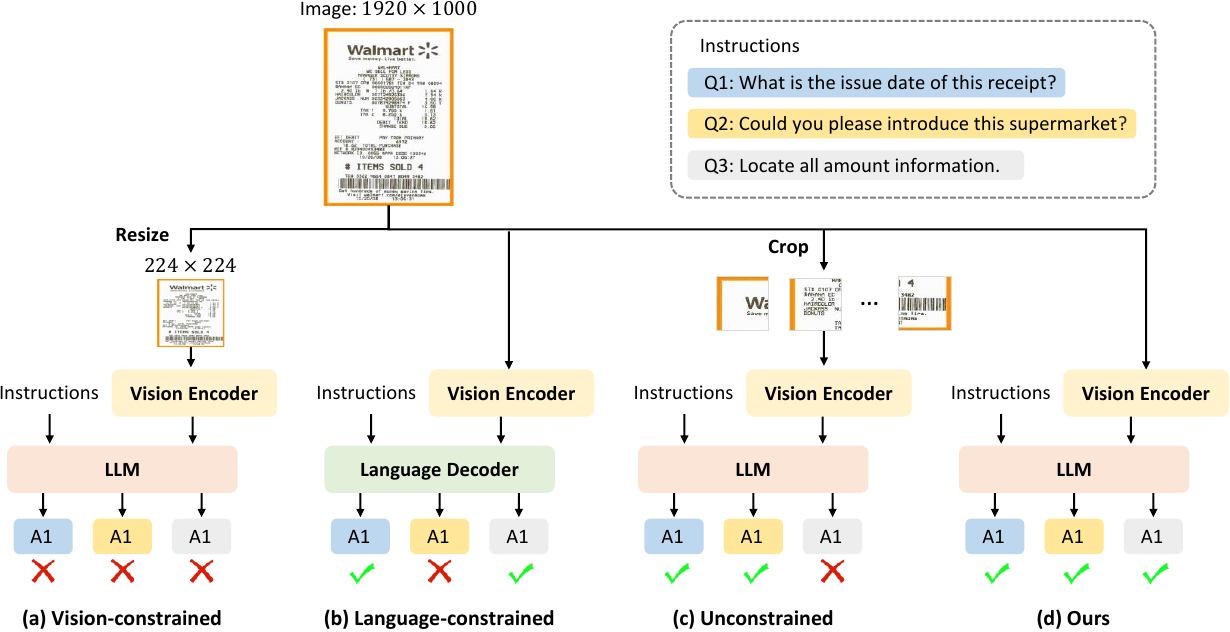}
	\caption{Comparisons of existing pipelines for document understanding. Contrasting with (a) vision-constrained, (b) language-constrained, and (c) unconstrained methods, our DocPedia efficiently processes high-resolution document images and performs logical reasoning using the world knowledge of large language models. The instructions Q1, Q2, and Q3 evaluate the text recognition, world knowledge, and text localization abilities, respectively.}     
       \label{fig1}
\end{figure}

Many early attempts~\cite{xu2021layoutxlm,xu2020layoutlm,huang2022layoutlmv3,hong2022bros,bai2022wukong,tang2023unifying,li2021structext,peng2022ernie,appalaraju2021docformer} in the field follow a perceive-then-comprehend paradigm, initially involving Optical Character Recognition~(OCR)~\cite{liao2020real,shi2016end} of document images, followed by the fusion of textual, layout, and visual features for content parsing.
However, the individual processing step of OCR may precipitate the accumulation of errors, leading to suboptimal performance. Furthermore, considering the intrinsic interweaving of visual elements and textual segments within documents, the reciprocity between perception and comprehension awaits further exploration. 

To attack this issue, OCR-free solutions have emerged as recent prevailing approaches in the field.
Among them, most methods commonly generate a sequence of tokens
that can be converted into a target string~\cite{ye2023mplug,feng2023unidoc,ye2023ureader,zhang2023llavar,ye2023mplug-doc} or a structured format data~\cite{kim2022ocr,lv2023kosmos,lee2023pix2struct}. 
Such generative models are skilled at synthesizing and rephrasing information, which naturally can unveil the implicit content or purpose behind the source material, as well as provide deeper insights and more versatile responses to the input inquiries. As depicted in Fig.~\ref{fig1}, they can be mainly categorized into three groups, namely (a) \textit{vision-constrained}, (b) \textit{language-constrained}, and (c) \textit{unconstrained} types, discussed next.

Specifically, in vision-constrained methodologies such as LLaVAR~\cite{zhang2023llavar}, mPLUG-DocOwl~\cite{ye2023mplug-doc}, and UniDoc~\cite{feng2023unidoc}, the visual encoders largely rely on a pre-trained CLIP-ViT~\cite{radford2021learning}, operating at input resolutions of 224 or 336. These resolutions are designed for images featuring texts in medium or large font sizes, \textit{e.g.}, scene text, but prove inadequate for text-intensive high-resolution documents where more details are indispensable~\cite{liu2023hidden}.
As shown in Fig.~\ref{fig1}~(a), when a high-resolution supermarket receipt is downscaled to 224 for model input, the text becomes unreadable, rendering these methods incapable of answering the three presented instructions.
In contrast, language-constrained approaches, including Donut~\cite{kim2022ocr}, KOSMOS-2.5~\cite{lv2023kosmos}, and Pix2Struct~\cite{lee2023pix2struct}, employ high-resolution input for training their models with a vision encoder. They abandon the use of large language models~(LLMs) in vision-constrained methods~\cite{zhang2023llavar,ye2023mplug-doc,feng2023unidoc}, and instead opt for a lightweight language decoder~\cite{vaswani2017attention}. While these approaches demonstrate promising perception ability, their comprehension performance is often compromised. This is because the vital components of robust logical reasoning and extensive world knowledge, typically provided by the LLM, are not adequately incorporated.
Taking Fig.~\ref{fig1}~(b) for example, in response to the instruction Q2, these models falter in providing accurate answers due to a deficiency in pertinent knowledge.

The \textit{status quo} triggers a question: \textit{Is there a feasible approach to maintain both perception and comprehension abilities without compromising vision and language?}   

To mitigate the problem in above both categories, unconstrained method~\cite{ye2023ureader}~(Fig.~\ref{fig1}~(c)) takes a further step by proposing a shape-adaptive cropping strategy. This strategy involves cropping high-resolution images into patches, which are then used in conjunction with a frozen low-resolution CLIP-ViT~\cite{radford2021learning} and LLM. However, this heuristic-based crop strategy may lead to semantic discontinuities, even after fusion is performed. Furthermore, the features extracted by CLIP-ViT~\cite{radford2021learning} are not well-suited for tasks that require fine-grained local detail, such as text detection~\cite{feng2023unidoc} or grounding (refer to Q3 in Fig.~\ref{fig1} (c)).

\begin{figure}[t]
	\centering
	\includegraphics[width=0.8\columnwidth]{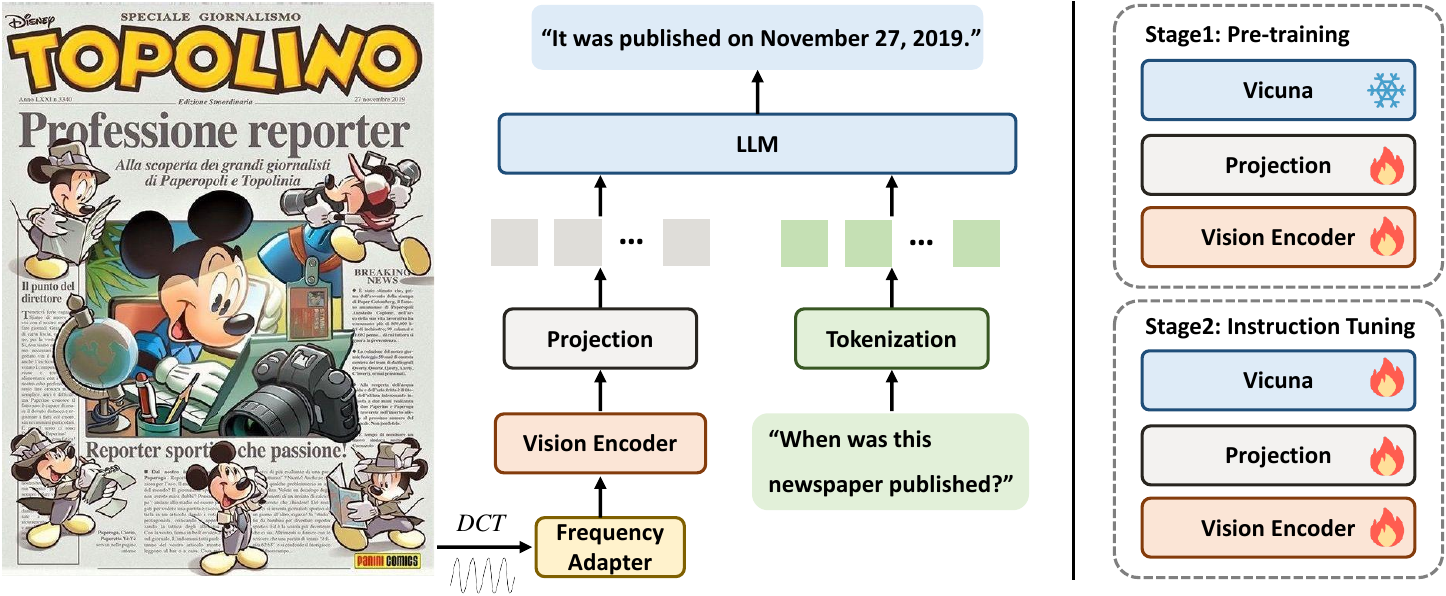}
	\caption{An overview of the proposed DocPedia. Given a document image, DocPedia initially applies JPEG DCT extraction to retrieve the DCT coefficients, which are then processed by a frequency adapter before feeding into a vision encoder. Then, the resultant tokens are combined with instruction-derived tokens, and fed into a large language model (LLM). Finally, based on these textual and visual information within the document as well as the intent of instruction, the LLM leverages its rich world knowledge to generate a logically coherent response.} 
	\label{fig:overview}
\end{figure}

To answer the question aforementioned, this work reinspects the problem through the lens of frequency and proposes DocPedia, an effective Large Multimodal Model (LMM), aiming to achieve versatile OCR-free document understanding.
DocPedia is capable of parsing high-resolution document images up to 2,560$\times$2,560, and harnessing the extensive world knowledge and powerful inference capabilities offered by LLMs~\cite{touvron2023llama,chiang2023vicuna}. This integration aims to enhance both perception and comprehension aspects. Technically, contrasting with previous LMMs in the filed, DocPedia directly processes visual input in the frequency domain~\cite{ahmed1974discrete,wallace1991jpeg,liu2023devil,liu2022nommer} rather than the pixel space. This unique characteristic enables DocPedia to capture a greater amount of visual and textual information using a limited number of visual tokens. 

Employing this effective architecture, we train our DocPedia with two phases: i)~\textit{text-aware pre-training} and ii)~\textit{context-aware fine-tuning}.
During pre-training, the vision encoder is trained to align the frequency domain features with a LLM~\cite{chiang2023vicuna}, incorporating various perception tasks across both document and natural scene contexts, such as text detection~\cite{liao2020real}, spotting~\cite{liu2018fots}, paragraph reading, image captioning~\cite{hossain2019comprehensive}, and \textit{etc}. 
In the subsequent fine-tuning stage, the focus shifts to the simultaneous learning of perception and comprehension, \textit{e.g.}, lower-level reading-related tasks, and higher-level document understanding. 
To ensure the robustness of the model as well as a consistent response style, we enrich the instructions and annotations of all these tasks with GPT~\cite{brown2020language}. Extensive quantitative and qualitative experiments are performed on this constructed large-scale instruction tuning dataset covering multiple document types. The results demonstrate the mutual benefits of jointly learning perception and comprehension tasks.

The contributions are summarized as follows:
\begin{itemize}
\item
To the best of our knowledge, we are the first to scale a large multimodal model for document understanding tasks to the resolution of 2,560$\times$2,560.
\item
We innovatively transform image domain inputs into frequency ones, enabling capturing more visual and textual information using a limited number of visual tokens.

\item
We achieved superior performance on multiple publicly available benchmark datasets and conducted extensive experiments to validate the effectiveness of DocPedia.
\end{itemize}

\section{Related work}
In the following, we provide an overview of existing research in the field of document understanding. The work is categorized into two distinct types: OCR-driven and OCR-free methodologies, discussed next.

\subsection{OCR-driven document understanding}
This section outlines methods that begin with text extraction from document images, followed by the integration of textual, layout, and visual features for thorough content analysis. Prominent among these are the LayoutLM series~\cite{xu2021layoutxlm,xu2020layoutlm,huang2022layoutlmv3}, which enhance text and layout modeling and integrate complex multimodal pre-training for richer representation learning. Wukong-Reader~\cite{bai2022wukong} employs pre-training objectives to exploit the structural knowledge of document textlines, incorporating textline-region contrastive learning for advanced visual document understanding. StrucTexT~\cite{li2021structext} combines a segment-token aligned encoder with diverse pre-training tasks, targeting enhanced structured text analysis in visually rich documents. DocFormer~\cite{appalaraju2021docformer} fuses text, vision, and spatial information using a distinct transformer architecture. However, the dependence of these methods on Optical Character Recognition (OCR) can result in error accumulation, raising efficacy concerns regarding the segregation of OCR in the context of the intertwined nature of visual and textual elements in documents.

\subsection{OCR-free document understanding}
To address this issue, prevailing OCR-free models excel in generating token sequences for varied responses and structured information synthesis, thereby offering enhanced insights and versatility in content creation and inquiry response.
Typically,
LLaVAR~\cite{zhang2023llavar} enhances document understanding by improving interaction skills with humans and boosting performance on text-rich image tasks, building upon its predecessor LLaVA~\cite{liu2023visual} with advanced visual instruction tuning techniques.
Based on the large multimodal model mPLUG-Owl~\cite{ye2023mplug}, mPLUG-DocOwl~\cite{ye2023mplug-doc} integrates a unified instruction tuning strategy across diverse document data.
UniDoc~\cite{feng2023unidoc} combines foundational OCR learning with text-rich image comprehension tasks, markedly boosting text scene image understanding.
Despite their strong representational skills and world knowledge from extensively pre-trained CLIP-ViT~\cite{radford2021learning} and large language models, these methods are limited to processing images with larger, sparser text due to the pre-trained visual models' lower resolution constraints.
StrucTexTv2~\cite{yu2023structextv2} employs self-supervised pre-training for document images, adeptly integrating masked image and language modeling~\cite{he2022masked,devlin2018bert} to enhance performance.
Donut~\cite{kim2022ocr} introduces an end-to-end trainable model that overcomes OCR limitations by using a synthetic document image generator for pre-training, enhancing performance in document understanding tasks.
Pix2Struct~\cite{lee2023pix2struct} through its unique pretraining on web page screenshots parsed into HTML, introduces a variable-resolution input and integrated language-vision approach.
As an evolution of Kosmos-2~\cite{peng2023kosmos}, Kosmos-2.5~\cite{lv2023kosmos} processes text-rich images, skillfully blending spatial text block creation and structured markdown generation in a streamlined, decoder-only model.
UReader~\cite{ye2023ureader} innovatively employs a shape-adaptive cropping module for high-resolution image processing. Monkey~\cite{li2023monkey} innovates LMMs with patch-based high-resolution processing and multi-level description, improving detail capture and context in visual-textual tasks. TextMonkey~\cite{liu2024textmonkey} innovates with Shifted Window Attention and token filtering, boosting performance across text-centric benchmarks and setting new standards in document understanding.

While these works exhibit outstanding outcomes in various aspects, many of them either struggle with handling high-resolution input or face challenges due to a lack of world knowledge. This underscores the future research endeavors: the development of an intelligent system adept at handling documents of various types and high resolutions.

\section{Method}
Fig.~\ref{fig:overview} presents an overview of DocPedia. 
It consists of two training phases:
(a) text-aware pre-training to align the visual features from the frequency domain to the large language model, and (b) context-aware fine-tuning for learning the parsing of documents. 
In the following, we first delineate the network architecture of DocPedia, followed by a detailed exposition of its two training phases.

\begin{figure}[t]
	\centering
	\includegraphics[width=0.7\columnwidth]{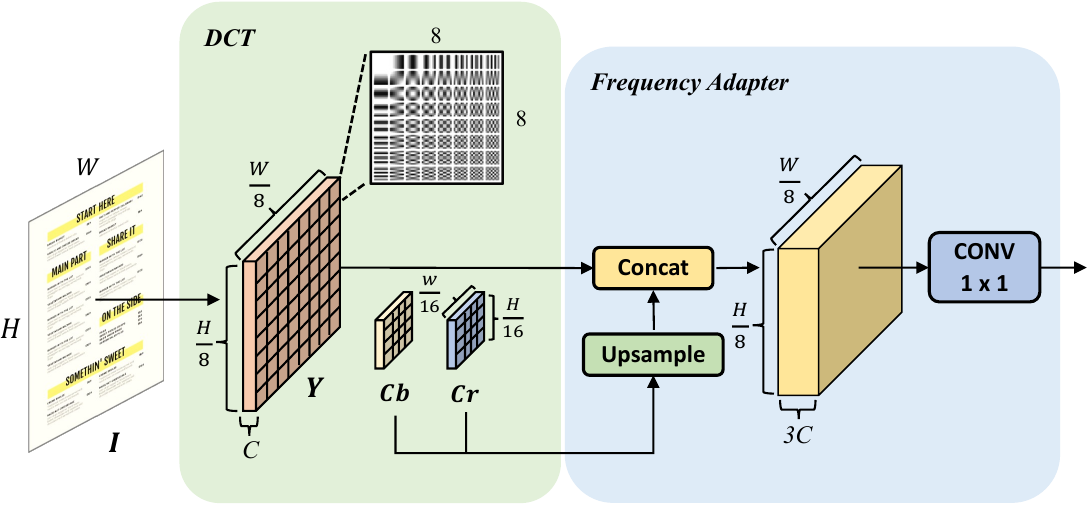}
	\caption{Schematic illustration of the DCT transformation and frequency adapter module in DocPedia.}     
	\label{fig:dct}
\end{figure}

\subsection{Architecture}
Given an input RGB document image, 
we first resize it to our designated training scale of $H\times W$ to obtain the image $\bm{I}$.
By default, both $H$ and $W$ are set as 2,560.
Here we preserve the aspect ratio during the resizing process to prevent distortion of textual elements.
Then, as shown in Fig.~\ref{fig:dct}, we apply the JPEG DCT extraction~\cite{ahmed1974discrete,wallace1991jpeg} to retrieve the DCT coefficients for the $\bm{Y}$, $\bm{Cb}$, and $\bm{Cr}$ channels.
The DCT coefficients are scaled down due to 8$\times$8 block processing for the luminance component ($\bm{Y}$) and additional chroma subsampling for color components ($\bm{Cb}$ and $\bm{Cr}$), resulting in $\frac{1}{8}$ and $\frac{1}{16}$ scales respectively.
Each of them features $C$ channels.
After that, we upscale $\bm{Cb}$ and $\bm{Cr}$ to a $\frac{1}{8}$ scale based on bilinear interpolation, followed by a concatenation along the channel dimension.
After this is a 1$\times$1 convolutional layer, employed to map the channel dimension of the concatenated map to that of the following backbone's input.
Through these operations, we acquire the frequency domain counterpart of image $\bm{I}$, denoted as $\bm{F}$.

Next, we feed $\bm{F}$ into the Swin Transformer~\cite{liu2021swin}, a visual backbone that leverages shifted windowing schemes to efficiently model spatial hierarchies. 
In our implementation, we remove the 1/4 scale downsampling module originally present before stage 1.
The output of the visual backbone is a feature map downsampled by a factor of 1/64. It is subsequently flattened, resulting in $\frac{H}{64}\times \frac{W}{64}$ tokens, each with a dimensionality of 1,024. Drawing inspiration from the paradigms of advanced large multimodal models~\cite{zhu2023minigpt,liu2023visual}, we employ a linear layer to align these tokens with the input token dimension of the following large language model~\cite{chiang2023vicuna}. The large language model employed in our DocPedia is Vicuna-7B~\cite{chiang2023vicuna}. It is derived from the LLaMA~\cite{touvron2023llama} and fine-tuned on curated conversational data to enhance its performance in dialogue and instruction-following tasks. Finally, the dimensionally aligned visual tokens are concatenated with the tokens transformed from the language instructions. This concatenated sequence is then fed into the LLM, generating the output response.

\subsection{Text-aware pre-training}\label{sec:pre}
To develop a vision encoder capable of processing frequency domain representation input and align it with the feature space of the following large language model~\cite{chiang2023vicuna}, we first undertook extensive text-aware pre-training.
During this stage, we freeze the large language model, focusing on the optimization of the vision encoder and its subsequent linear projector, as illustrated in Fig.~\ref{fig:overview}.

Specifically, our pre-training encompassed a variety of perception tasks, including text detection~\cite{liao2020real}, recognition~\cite{wang2011end}, spotting~\cite{liu2018fots}, paragraph reading, full-text reading~\cite{kim2022ocr}, and image captioning~\cite{hossain2019comprehensive}. The first three tasks are foundational OCR tasks. ``Paragraph reading" denotes the reading of the text within a specified bounding box (see bottom case in Fig.~\ref{fig:demo_perc}), whereas ``full-text reading" refers to deciphering all text in the image. It is worth noting that the first five tasks focus on a diverse array of document images, while the final task targets natural scene images. This comprehensive pre-training enables the vision encoder of our DocPedia to effectively perceive textual and visual information from both document and natural scene images.

\begin{table}[t]
\footnotesize
\centering
\renewcommand{\arraystretch}{1.28}
\setlength{\tabcolsep}{0.6mm}
\caption{Summary of the training data statistics across two stages. The symbols represent various instruction-following tasks as follows: $\mathcal{D}$ for text detection, $\mathcal{R}$ for text recognition, $\mathcal{S}$ for text spotting, $\mathcal{R}_p$ for paragraph reading, $\mathcal{R}_f$ for full-text reading, $\mathcal{C}$ for image captioning, and $\mathcal{U}$ for document understanding. \textbf{\# Conv} denotes the number of the samples.}
\begin{tabular}{cccccc}
\toprule
\textbf{Stage} & \textbf{Image} & \textbf{Instruction} & \textbf{Task} & \textbf{\# Conv} \\ 
\midrule
\multirow{3}{*}{Pre-training} & Scene & LLaVA~\cite{liu2023visual} & $\mathcal{C}$ & 595K \\
 & PDF & OCR & $\mathcal{D},\mathcal{R},\mathcal{S},\mathcal{R}_p,\mathcal{R}_f$ & 325K \\
 & PPT & OCR & $\mathcal{D},\mathcal{R},\mathcal{S},\mathcal{R}_p,\mathcal{R}_f$ & 600K \\
\midrule
\multirow{4}{*}{Fine-tuning} 
 & PDF & OCR & $\mathcal{D},\mathcal{R},\mathcal{S},\mathcal{R}_p,\mathcal{R}_f$ & 325K \\
 & PPT & OCR & $\mathcal{D},\mathcal{R},\mathcal{S},\mathcal{R}_p,\mathcal{R}_f$ & 600K \\
& Scene & LLaVA~\cite{liu2023visual} & $\mathcal{U}$ & 158K \\
 & Benchmark & GPT & $\mathcal{U}$ & 370K \\
\bottomrule
\end{tabular}
\label{table:dataset-summary}
\end{table}

\begin{table}[t]
\footnotesize
\caption{Different types of OCR instructions and their examples.}
\centering
\begin{tabular}{>{\centering\arraybackslash}m{3cm}|m{6.5cm}}
\toprule
\textbf{Type} & \multicolumn{1}{c}{\textbf{Example}} \\
\hline
Detection & ``Where are the texts located in the photo?"  \\
\hline
Recognition &  ``Recognize all the text in this image." \\
\hline
Spotting & ``Identify all the text in the shot and return their coordinates in the format of [x1,y1,x2,y2]." \\
\hline
Paragraph Reading & ``Tell me about the content in the area marked as [0.124,0.276,0.353,0.487] of the frame." \\
\hline
Full Text Reading & ``Convey the entire content of this pic to me." \\
\bottomrule
\end{tabular}
\label{tab:vision_commands}
\end{table}

\begin{table*}[t]
    \centering
\caption{Quantitative accuracy ($\%$) comparison with existing large multimodal models (LMMs) on key information extraction (KIE) and visual question answering (VQA) benchmarks.
``Supervised-SOTA" reported in~\cite{liu2023hidden} denotes specialized models that have been fine-tuned on the training set of the corresponding benchmark.
Black bold highlights represent the highest performance, while underlined highlights denote the second-best performance.}
    \resizebox{1\linewidth}{!}{\begin{tabular}{c|c|ccc|ccc|ccc|c}
    \toprule      \multirow{2}{*}{Method} &  \multirow{2}{*}{Resolution}
                 & \multicolumn{3}{c|}{Scene Text-Centric VQA}        & \multicolumn{3}{c|}{Document-Oriented VQA}                    & \multicolumn{3}{c|}{KIE}  & \multirow{2}{*}{OCRBench}    \\
                 & & STVQA & TextVQA & OCRVQA  & DocVQA & InfoVQA & ChartQA & FUNSD   & SROIE  & POIE  \\ \midrule
    BLIP2-OPT-6.7B~\cite{li2023blip} & 224$\times$224   & 20.9  & 23.5    & 9.7            & 3.2    & 11.3           & 3.4                  & 0.2     & 0.1    & 0.3     & 235    \\
    mPLUG-Owl~\cite{ye2023mplug}  & 224$\times$ 224 & 30.5  & 34.0      & 21.1           & 7.4    & 20.0             & 7.9           & 0.5     & 1.7    & 2.5   & 297   \\
    InstructBLIP~\cite{dai2024instructblip} & 224$\times$ 224 & 27.4  & 29.1    & 41.3        & 4.5    & 16.4           & 5.3             & 0.2     & 0.6    & 1.0   & 276       \\
    LLaVAR~\cite{zhang2023llavar}   & 224$\times$224    & 39.2  & 41.8    & 24.0             & 12.3   & 16.5           & 12.2              & 0.5     & 5.2    & 5.9    & 346    \\
    BLIVA~\cite{hu2024bliva}    & 224$\times$224     & 32.1  & 33.3    & 50.7      & 5.8    & \underline{23.6}           & 8.7             & 0.2     & 0.7    & 2.1  & 291     \\
    LLaVA1.5-7B~\cite{liu2023improved} & 336$\times$336    & 38.1  & 38.7    & 58.1         & 8.5    & 14.7           & 9.3            & 0.2     & 1.7    & 2.5   & 297       \\
    TGDoc~\cite{wang2023towards} & 336$\times$336   & 36.3 & 46.2    & 37.2     & 9.0           & 12.8                & 12.7       & 1.4    & 3.0   & 2.2  & - \\
    UniDoc~\cite{feng2023unidoc}  & 336$\times$336     & 35.2  & 46.2    & 36.8           & 7.7    & 14.7           & 10.9                    & 1.0       & 2.9    & 5.1  & -      \\
    mPLUG-Owl2~\cite{ye2023mplug2}  & 448$\times$448  & 49.8  & 53.9    & 58.7        & 17.9   & 18.9           & 19.4            & 1.4     & 3.2    & 9.9      & 366  \\
    InternLM-XComposer2~\cite{dong2024internlm}    & 490$\times$490   & \textbf{59.6} & \underline{62.2}    & 49.6           &39.7    & \textbf{28.6}           & \underline{51.6}  & 15.3   & 34.2    & \textbf{49.3}  & 511 \\ 
    Monkey~\cite{li2023monkey}    & 1,344$\times$896    & \underline{54.7}  & \textbf{64.3}    & \textbf{64.4}    & \textbf{50.1}   & \underline{25.8}           & \textbf{54.0 }           & 24.1    & {41.9}   & 19.9   & 514  \\ 
    \midrule
    DocPedia  & 1,920$\times$1,920     & 43.2  & 54.0    & 43.7         & 39.5   & 15.2           & 41.6                 & \underline{30.5}   & \underline{53.6}  & 43.5   & 476    \\
    DocPedia$^\dagger$   & 2,560$\times$2,560   & 47.3  & 61.2   & \underline{59.9}         & \underline{49.3}   & 15.5           & 47.8             &  \textbf{40.1}  & \textbf{57.7}   & \underline{48.8}    & 507  \\
    \bottomrule
    \end{tabular}}
    \label{tab:per_com}
\end{table*}

\subsection{Context-aware fine-tuning}
During fine-tuning, we concurrently cultivate the perception and comprehension capabilities of DocPedia. Concretely, within each batch of training data, one half is dedicated to the five types of OCR tasks outlined in the pre-training phase, while the other half comprises tasks that demand a higher level of semantic understanding related to document~\cite{mathew2021docvqa} and scene~\cite{liu2023visual}.
We argue that the concurrent learning of lower-level perceptual abilities and the cultivation of higher-level understanding capabilities can maximize the performance of the model. During this stage, we unfreeze the LLM and fine-tune the entire model.

\section{Dataset construction}                                           
To train our DocPedia,
we construct a large-scale multimodal instruction following dataset.
The statistical data employed during the pre-training and fine-tuning phases are summarized in Table~\ref{table:dataset-summary},
introduced next.

\subsection{Pre-training}
During the pre-training phase, our focus was on the learning of perceptual abilities, particularly in the context of text perception. 
As illustrated in Table 1, we amassed a dataset comprising 600,000 PowerPoint (PPT) images and 325,000 PDF images.
The PowerPoint images are sourced from the ``Common Crawl" dataset\footnote{https://commoncrawl.org/}, an extensive web corpus encompassing publicly accessible web pages. The PDF images are sourced from arXiv\footnote{https://arxiv.org/}, an established online platform for scientists to publish pre-print research papers.

For each of these images, we randomly selected an Optical Character Recognition (OCR) task type as described in Sec.~\ref{sec:pre} and then constructed corresponding instructions and responses~\cite{feng2023unidoc}. On one hand, to ensure instruction diversity, we generated multiple variations of instructions for each OCR task using GPT-3.5~\cite{brown2020language}. In Table~\ref{tab:vision_commands}, we present one exemplar for each of the five text-aware perceptual tasks.
For further examples, please refer to the supplementary materials. On the other hand, for their responses, we employed a standardized format (see Fig.~\ref{fig:demo_perc}). In addition to the aforementioned data, we enriched our dataset with 595,000 caption entries from LLaVA~\cite{liu2023visual}, aiming to enhance the DocPedia's perceptual abilities in natural scenes.

\subsection{Fine-tuning}
Furthermore, during the fine-tuning phase,
we first employed the same data utilized during the pre-training phase, comprising 325,000 PDF and 600,000 PPT images.
Building upon this, we introduced an extra 370,000 entries from seven visual question answering benchmark datasets, including DocVQA~\cite{mathew2021docvqa}, OCRVQA~\cite{mishra2019ocr}, TextVQA~\cite{singh2019towards}, InfoVQA~\cite{mathew2022infographicvqa}, ChartVQA~\cite{masry2022chartqa}, FigureVQA~\cite{kahou2017figureqa}, PlotVQA, FUNSD~\cite{jaume2019funsd}, SROIE~\cite{huang2019icdar2019}, and POIE~\cite{kuang2023visual}. 
Notably, as the responses in these datasets are typically concise, containing only the answer itself, we employed GPT-3.5~\cite{brown2020language} to expand these responses into complete sentences. This adaptation was done to align with the characteristic comprehensive and detailed response style of large language models~\cite{chiang2023vicuna}.
Besides, we supplemented the training data with 158,000 instruction tuning data for natural scene understanding from LLaVA~\cite{liu2023visual}. Our experiments demonstrate the effectiveness of a fine-tuning strategy that concurrently learns perceptual and understanding abilities.

\section{Experiment}
\subsection{Implementation details}
To implement DocPedia, we adopted a one-cycle learning rate strategy~\cite{smith2019super}. 
For the pre-training phase, the peak learning rate was established at 1e-3, which was set as 1e-5 during the subsequent fine-tuning phase.
We maintained batch sizes of 64 and 8 for the pre-training and fine-tuning stages, respectively. We employ the AdamW optimizer~\cite{loshchilov2017decoupled} and both training stages were performed on eight A100 GPUs, each spanning just a single epoch.

For performance assessment, a temperature parameter of 0.2 was utilized in both quantitative and qualitative evaluations. 
We adopted the accuracy metric, where a response generated by the model is considered correct if it contains the string present in the ground truth~\cite{liu2023hidden}.

\begin{figure}[t]
	\centering
	\includegraphics[width=0.6\columnwidth]{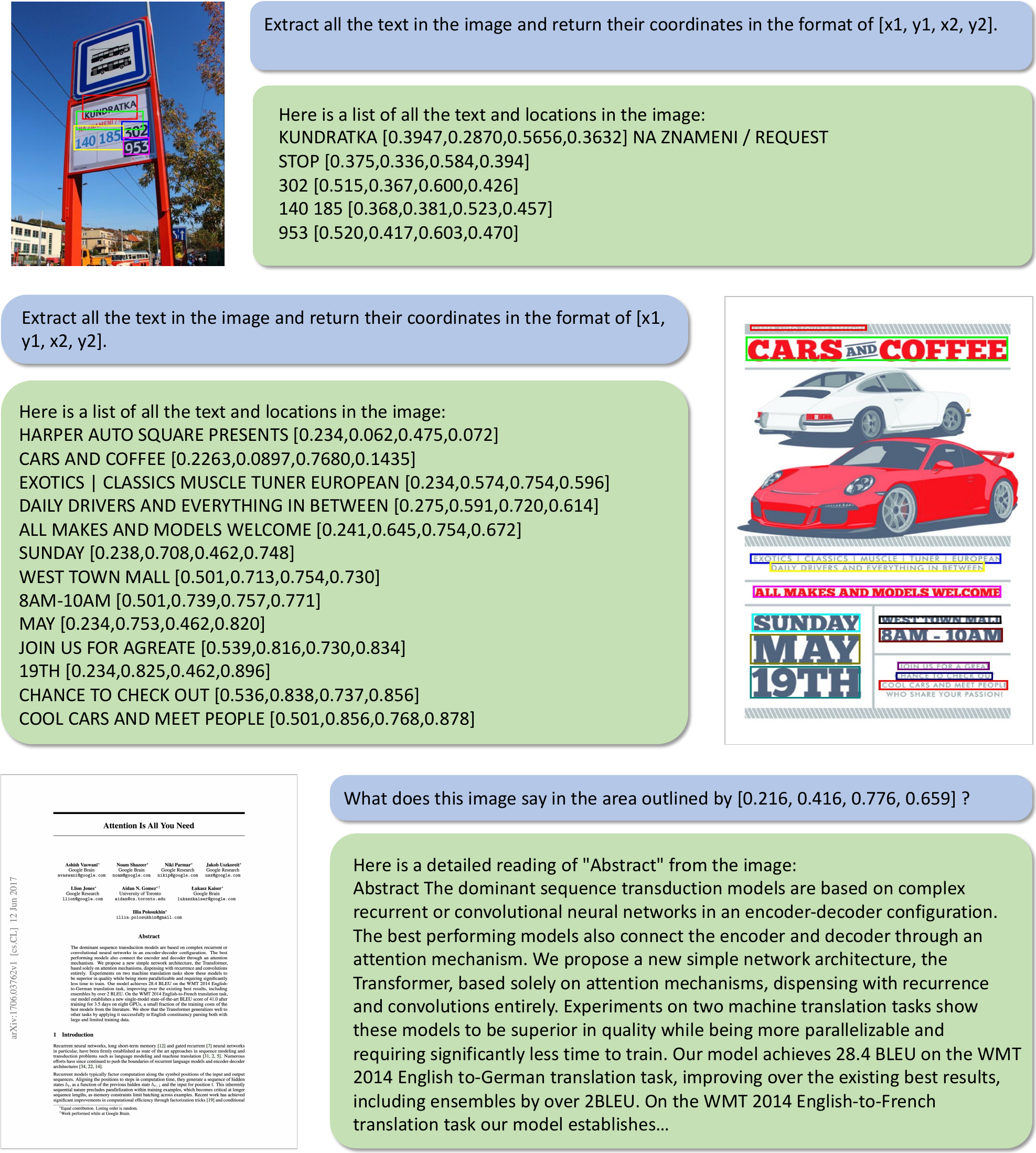}
	\caption{Exemplary demonstrations of DocPedia's advanced text perception capabilities. The three instances illustrate its adeptness in accurately identifying and localizing text in scene and document images, and demonstrating proficient paragraph reading skills. We visualized the bounding boxes within the responses in the images. For the last case, subsequent text readouts have been omitted for display convenience. Zoom in for best view.}     
	\label{fig:demo_perc}
\end{figure}

\begin{figure}[h]
	\centering
	\includegraphics[width=0.92\columnwidth]{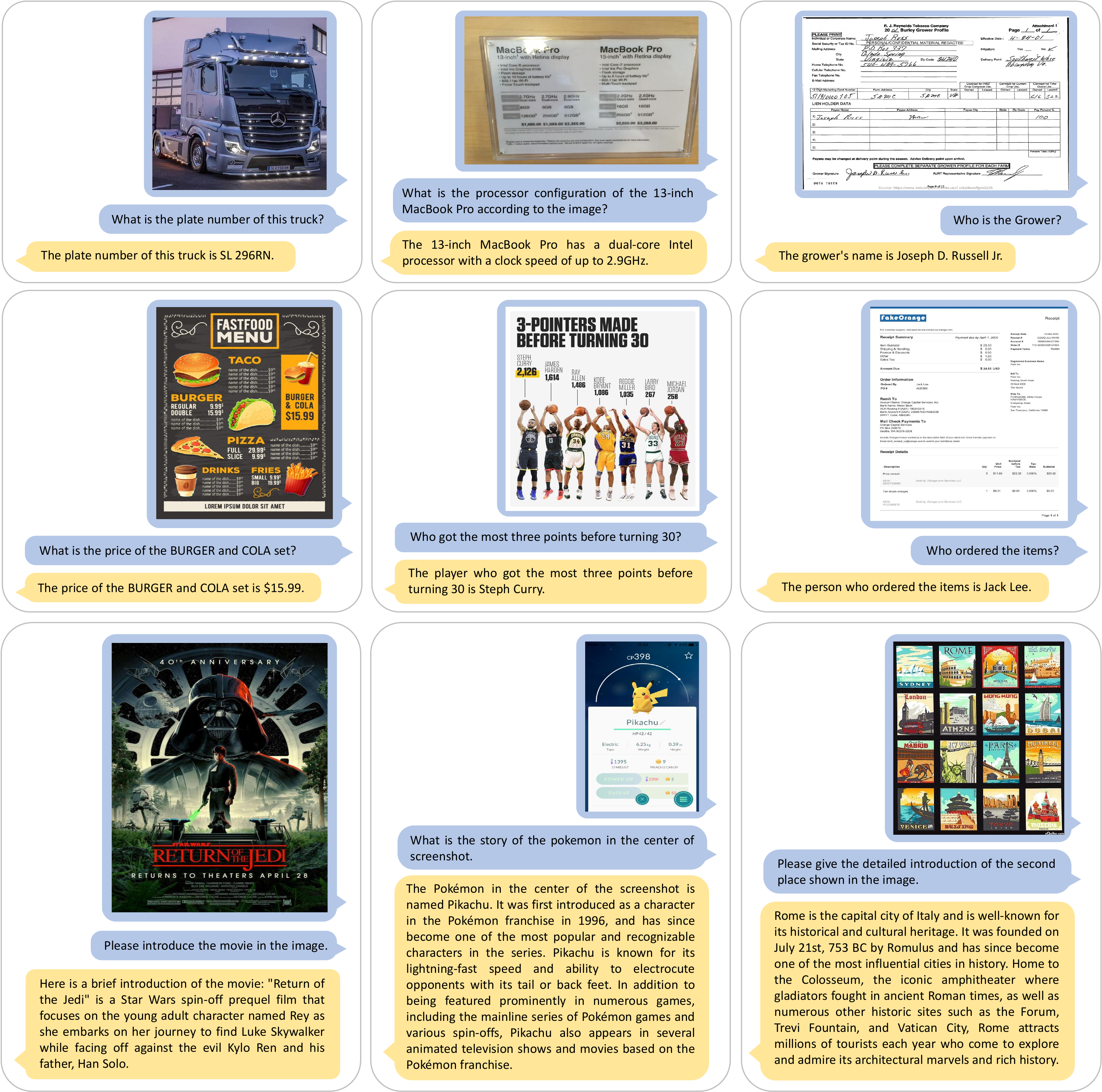}
	\caption{Exemplary demonstrations of DocPedia’s comprehension capabilities. The first two rows primarily showcase DocPedia’s accurate responses to the instructions based on visual and textual information within images. Furthermore, the bottom row further exhibits DocPedia’s ability to leverage its extensive world knowledge from its large language model component for responding. The evaluated images include high-resolution textual scene images and document images. Best viewed in color and zoom in.}     
	\label{fig:demo}
\end{figure}

\begin{figure}[h]
	\centering
	\includegraphics[width=0.93\columnwidth]{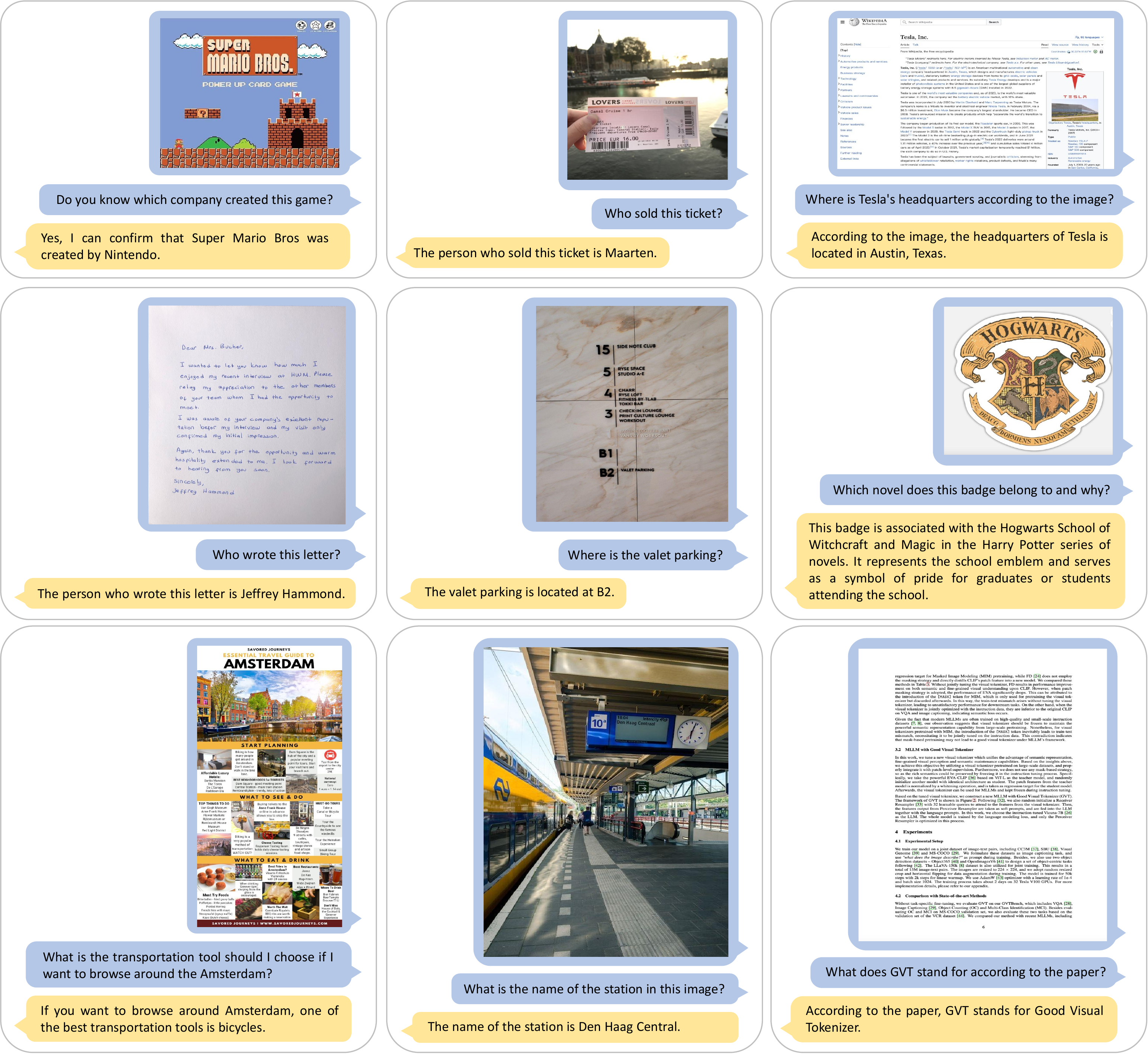}
	\caption{Exemplary demonstrations of DocPedia’s comprehension capabilities. DocPedia accurately responds to instructions by analyzing both visual and textual information within images. It exhibits effective proficiency in interpreting scene text images and document images. Notably, it exhibits robust performance in processing images containing text of varying sizes, fonts, and styles.}     
	\label{fig:demo}
\end{figure}

\subsection{Results}
We further conducted both quantitative and qualitative evaluations of the current state-of-the-art multimodal large-scale models in comparison to our proposed method.

\textbf{Qualitative results.}
We qualitatively evaluate DocPedia's perception and comprehension capabilities on high-resolution scene text and document images.
Firstly, in terms of the perception capabilities, as illustrated in Fig.~\ref{fig:demo_perc}, our DocPedia can accurately locate and identify text in both scenes and high-resolution documents, which is attributed to the training of fundamental OCR tasks in Table~\ref{table:dataset-summary}. 
Secondly, regarding comprehension abilities,
as demonstrated in Fig.~\ref{fig:demo}, the examples in the first two rows indicate that DocPedia can perceive and understand the visual and textual information in images to provide accurate responses, based on the intention of the instructions. Moreover, the examples in the bottom row illustrate that DocPedia is capable of integrating the content of instructions, visual and textual information within images, and its large language model's rich world knowledge to formulate responses.
These results demonstrate DocPedia's robust multimodal comprehension capabilities. For additional examples, please refer to the supplementary materials.

\begin{table}[t]
\footnotesize
\centering
\caption{Ablation experiments regarding the use of various resolutions in the RGB domain and frequency domain as inputs for the vision encoder in DocPedia. ``Tokens" refers to the number of tokens outputted by the vision encoder.}
\begin{tabular}{ccccc} 
   \toprule
   \multicolumn{3}{c}{Method} & \multicolumn{2}{c}{VQA}  \\
   \cmidrule(lr){1-3}\cmidrule(lr){4-5}
   Input & Resolution & Tokens & DocVQA~\cite{mathew2021docvqa} & TextVQA~\cite{singh2019towards}  \\
   \midrule
   RGB & 640$\times$640 & 400 & 15.1 & 28.9 \\
   RGB & 960$\times$960 & 900 & 22.5 & 42.3 \\
   RGB & 1,280$\times$1,280 & 1,600 & 30.1 & 49.2 \\
   \midrule
   DCT & 1,280$\times$1,280 & 400 & 22.4 & 46.3 \\ 
   DCT & 1,920$\times$1,920 & 900 & 39.5 & 54.0 \\
   DCT & 2,560$\times$2,560 & 1,600 & \textbf{49.3} & \textbf{61.2} \\
   \midrule
   RGB Flattening  & 1,920$\times$1,920 & 900 & 31.92 & 47.01 \\
   RGB Flattening & 2,560$\times$2,560 & 1,600 & 41.08 & 52.18 \\
   \bottomrule
\end{tabular}
\label{aba1}
\end{table} 

\begin{table}[t]
\footnotesize
\caption{Ablation experiments concerning the training strategies of DocPedia during the pre-training and fine-tuning phases. All ablations are conducted at a resolution of 1,920$\times$1,920.}
\centering
\begin{tabular}{c|c|ccc|cc} 
   \toprule
    & \multirow{2}{*}{Pre-training} & \multicolumn{3}{c|}{Fine-tuning} & \multicolumn{2}{c}{VQA}  \\
   \cmidrule(lr){3-4}\cmidrule(lr){5-5}\cmidrule(lr){6-7}
   & & Text Perception & Position Perception & Understanding & DocVQA~\cite{mathew2021docvqa} & TextVQA~\cite{singh2019towards} \\
   \midrule
    (a) &           & \checkmark & \checkmark & \checkmark & 23.0 & 35.5 \\
    (b) & \checkmark &            &            & \checkmark & 28.7 & 49.6 \\
    (c) & \checkmark & \checkmark &            & \checkmark & 35.8 & 52.9 \\
    (d) & \checkmark & \checkmark & \checkmark & \checkmark & \textbf{39.5} & \textbf{54.0} \\
   \bottomrule
\end{tabular}
\label{aba2}
\end{table}

\begin{figure}[t]
	\centering

	\includegraphics[width=0.9\columnwidth,height=0.22\columnwidth]{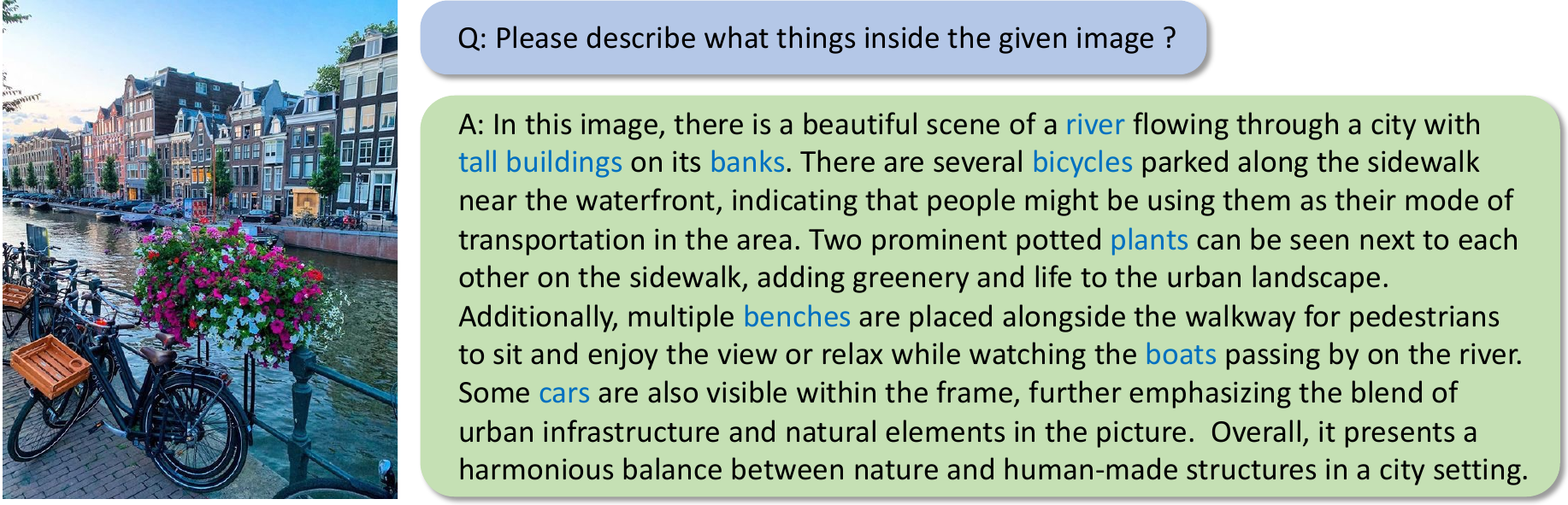}
         \caption{Exemplary demonstration of DocPedia's transferability on high-resolution general scene images.}
         \label{fig:transfer}
\end{figure}

The main focus of our DocPedia is to explore high-resolution document understanding and the relationship between perception and understanding abilities. 
Hence, we further confirm the transferability of our method's high-resolution perception ability for general scene understanding.
As shown in Fig.~\ref{fig:transfer}, our DocPedia can accurately detect small-scale car objects.

\textbf{Quantitative results.}
Furthermore, we conduct a comprehensive quantitative evaluation for the existing large multimodal models and our DocPedia. The results are summarized in Table~\ref{tab:per_com}. The benchmarks used for this assessment consist of 3 scene text-centric VQA benchmark datasets, including STVQA~\cite{biten2019icdar}, TextVQA~\cite{singh2019towards}, and OCRVQA~\cite{mishra2019ocr}, 3 document-oriented VQA benchmark datasets, including DocVQA~\cite{mathew2021docvqa}, ChartVQA~\cite{masry2022chartqa}, InfoVQA~\cite{mathew2022infographicvqa}, and 3 Key Information Extraction (KIE) benchmark datasets, including FUNSD~\cite{jaume2019funsd}, SROIE~\cite{huang2019icdar2019} as well as  POIE~\cite{kuang2023visual}. 

As we can see, on several high-resolution document image benchmarks~\cite{jaume2019funsd,huang2019icdar2019,kuang2023visual,mathew2021docvqa,masry2022chartqa}, where the text is dense and tiny, our DocPedia demonstrates significant performance improvements over existing advanced large multimodal models. Notably, compared with the state-of-the-art method Monkey~\cite{li2023monkey}, our DocPedia achieved comparable performance on the challenging DocVQA~\cite{mathew2021docvqa} benchmark dataset. On the FUNSD~\cite{jaume2019funsd} and SROIE~\cite{huang2019icdar2019} benchmark dataset, our DocPedia outperformed Monkey~\cite{li2023monkey} by 16.0\% and 15.8\%, respectively.
These results underscore the distinct advantages of our approach. Besides, our DocPedia also performed well on OCRBench~\cite{liu2023hidden}, a comprehensive benchmark with 29 OCR-related evaluations. It achieved a score of 507, which represents an improvement over previous large-scale multimodal models for document understanding.

However, it is crucial to note that in the domain of multimodal document understanding, overall performance is not solely contingent upon the resolution of the input image. Additional influential factors include the quantity and quality of instruction tuning data, the type of visual encoder employed, the specific large language model utilized and so on. Specifically, our DocPedia re-trains the visual encoder and utilizes Vicuna-7B~\cite{chiang2023vicuna} as the LLM component. In contrast, the advanced Monkey~\cite{li2023monkey} performs direct instruction tuning on the well-trained Vit-BigG and the LLM from the large multimodal model Qwen-VL~\cite{Qwen-VL}.

\subsection{Ablation studies}
We further conduct extensive ablation studies to validate the efficacy of core settings and components in our DocPedia.
Note that all experiments were conducted on two benchmark datasets: 
DocVQA~\cite{mathew2021docvqa} and TextVQA~\cite{singh2019towards}.
DocVQA~\cite{mathew2021docvqa} is centered around document comprehension, whereas TextVQA~\cite{singh2019towards} focuses on scene text image understanding. 
Both datasets are notable for their substantial sample sizes, comprising 5,000 and 5,349 test samples, respectively.

\begin{figure}[t]
	\centering
	\includegraphics[width=0.65\columnwidth]{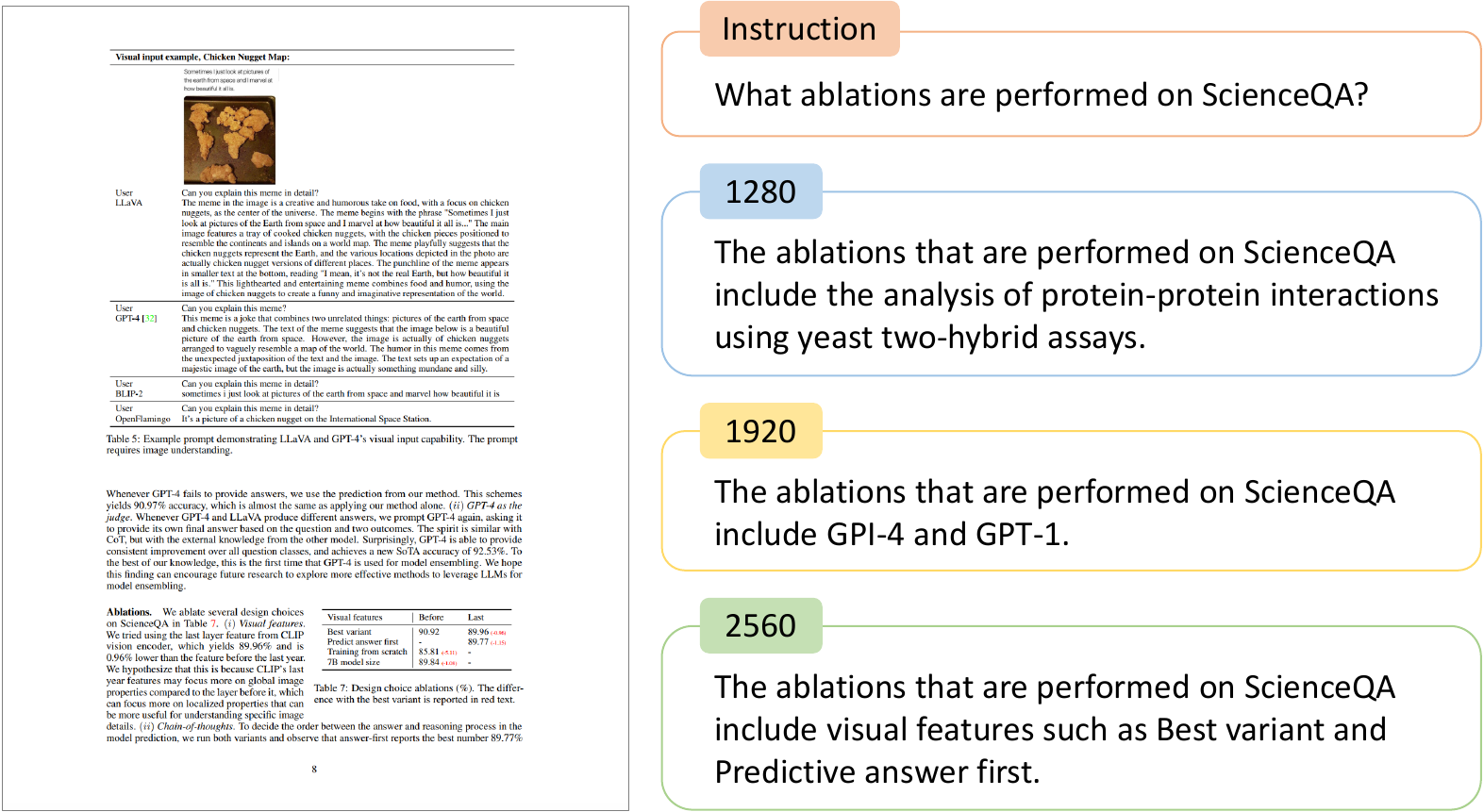}
	\caption{Comparison of DocPedia's responses to varying resolutions of DCT inputs for the same high-resolution document image, encompassing scales of 1,280, 1,920, and 2,560. The response becomes accurate at a scale of 2,560. Zoom in for the best view.}     
	\label{fig:reso_aba}
\end{figure}

\begin{figure*}[t]
	\centering
	\includegraphics[width=1\columnwidth]{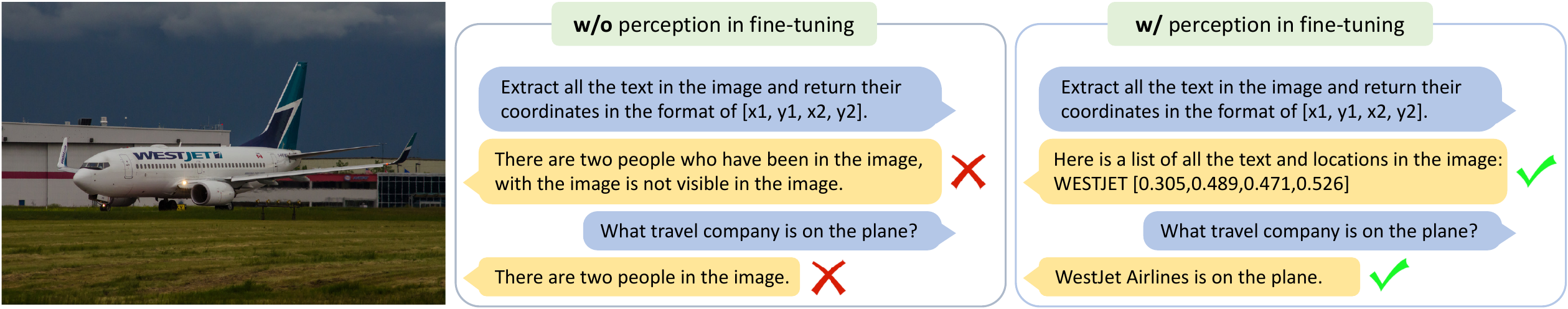}
	\caption{Illustration demonstrating the impact of \textbf{joint learning of perception and comprehension} during fine-tuning. We present a comparison between two models under the same conditions: a given image with identical spotting and understanding instructions. The model fine-tuned solely on the understanding task (middle) responds incorrectly to both instructions, whereas the model undergoing joint learning of perception and understanding tasks (right) during fine-tuning answers both tasks correctly.} 
	\label{fig:supp_aba1}
\end{figure*}

\textbf{Impact of training in the frequency domain.}
One of the significant contributions of our DocPedia lies in utilizing the frequency domain representation of images as the input for the vision encoder.
In Table~\ref{aba1}, we first evaluate our method's performance using RGB inputs and frequency domain inputs on varying scales. For RGB inputs, three resolution settings were evaluated: 640, 960, and 1,280. Given that the backbone Swin Transformer~\cite{liu2021swin} downsamples the input image by a factor of 32, the resultant token counts are 400, 900, and 1,600, respectively. In experiments with our frequency domain inputs, we tested image resolutions of 1,280, 1,920, and 2,560 for the DCT, resulting in token counts corresponding to the three image-based experimental settings.

As we can see, with the same number of visual tokens, our DocPedia yields better performance. This is attributed to the increased resolution enabling enhanced perception of texture content within images. In experiments where the input resolution is consistent (1,280 in Table~\ref{aba1}), we observe a slightly enhanced performance with image inputs compared to frequency ones. 
Note that the number of visual tokens for the latter is only a quarter of that used for the former.
This is likely because our frequency-based approach retains a limited number of tokens, leading to some information loss. However, this constraint simultaneously facilitates the incorporation of higher-resolution inputs, up to 2,560$\times$2,560.
In Fig.~\ref{fig:reso_aba}, we further compare the responses of DocPedia to the same academic image and instruction under varying input resolutions. It is observed that the response becomes accurate when input resolution reaches 2,560.

To further investigate the necessity of frequency learning in enhancing input image resolution, we conducted comparative experiments involving different input processing techniques (RGB Flattening vs. DCT), as outlined in Table~\ref{aba1}. Specifically, in the ``RGB Flattening" approach, we initially divided the input image ($H \times W \times 3$) into non-overlapping patches of size $8\times 8 \times 3$ without applying the DCT. Subsequently, each patch was directly unfolded into a $1\times 1 \times 192$ vector, forming a resulting cube of size $\frac{H}{8} \times \frac{W}{8} \times 192$, which shares a similar shape with the counterpart in our ``DCT" approach.

From the observations presented in Table~\ref{aba1}, it is evident that employing the ``DCT" method consistently yields significantly higher performance, whether with a resolution of 1,920 or 2,560, compared to the simplistic ``RGB Flattening" technique. Moreover, we surprisingly find that introducing DCT leads to more stable and faster convergence during training, particularly when dealing with high-resolution document images. We attribute these phenomena to the unique characteristics of DCT. As an orthogonal transformation, DCT decouples and orthogonalizes frequency components, effectively reducing input redundancy. In contrast, direct flattening in the RGB space lacks this capability. Although it can be partially compensated for through learning from large-scale data using convolutional or linear layers~\cite{dosovitskiy2020image}, it is evident that they do not achieve the same level of direct redundancy reduction as the DCT.

\textbf{Impact of the training strategy.}
In Table~\ref{aba2}, we further study the impact of our training strategies. Initially, we omitted the pre-training phase (experiment (a)), opting instead for a random initialization of the vision encoder. Significant performance degradation was observed in the absence of pre-training, underscoring the critical role of feature alignment between the vision encoder and subsequent LLM~\cite{chiang2023vicuna}.

Furthermore, we examined the fine-tuning strategies. Under default settings, we concurrently learn perceptual and understanding capabilities, incorporating tasks OCR, image captioning, document understanding, and scene comprehension.
Hence, we first conducted ablation studies focusing on the perception tasks. 
Here, the perception tasks can be categorized into two primary domains: position perception and text perception. Position perception tasks encompass text detection and spotting.
Text perception tasks comprise text recognition, paragraph reading, and full text reading. 
The results are presented in Table~\ref{aba2}.
The experiments (b), (c), and (d) compare three settings during the fine-tuning phase: (b) no perception tasks trained, (c) training only text perception tasks, and (d) incorporating all perception tasks. The results demonstrate that the inclusion of position perception tasks led to improved performance, underscoring their significance in text-rich image understanding. Similar conclusions were experimentally validated in UniDoc~\cite{feng2023unidoc}, where the performance of text recognition is improved when text detection tasks are incorporated.
The results of experiment (b) and (d) affirming the efficacy of our joint training strategy. This implies that the simultaneous development of foundational perceptual skills augments the acquisition of comprehension abilities.

Furthermore, we provide a vivid example in Fig.~\ref{fig:supp_aba1} to confirm the facilitation of perception to comprehension. Concretely, we compare the responses of two models under identical conditions: the same image and instructions for text spotting and image understanding. 
The model~(``w/o perception in fine-tuning''), which is fine-tuned merely with the understanding task (middle), fails to recognize the target text on the plane and responds the problematic answer to the understanding question. In contrast, the other model~(``w/ perception in fine-tuning'') fine-tuned on both perception and understanding tasks, accurately responds to the two instructions.

\subsection{Limitation discussion}
Furthermore, we discuss the limitations of our DocPedia. Firstly, as illustrated in Table~\ref{tab:per_com}, we observe minimal performance improvements on the InfoVQA dataset~\cite{mathew2022infographicvqa}. This highlights one of the constraints of DocPedia. Many images in InfoVQA~\cite{mathew2022infographicvqa} possess extremely high aspect ratios, akin to vertically concatenating multiple pages of images, with some even reaching dimensions of 6,400$\times$800.
In addition, our DocPedia currently lacks the capability to process multi-page document images~\cite{tito2023hierarchical} and also exhibits a deficiency in multilingual proficiency~\cite{Qwen-VL}.

\section{Conclusion}
This work introduces DocPedia, an innovative Large Multimodal Model tailored for versatile OCR-free document understanding, capable of handling images with high resolutions. Unlike existing methods, DocPedia directly processes visual input in the frequency domain, where more visual and textual information is captured in a limited number of visual tokens. Thanks to the dual-stage training strategy designed and the polished instructions/annotations for all tasks, DocPedia shows superior performance on several public datasets. In conclusion, we provide a successful attempt at pathways for handling complex high-resolution documents. We expect our success in exploring LMM dealing with high-resolution images from frequency perspective could trigger more insights for the community.

\bibliographystyle{scis}
\bibliography{ebbib}

\begin{thebibliography}{10}
\providecommand{\url}[1]{\texttt{#1}}
\providecommand{\urlprefix}{URL }
\providecommand{\bibinfo}[2]{#2}

\bibitem{srihari1986document}
\bibinfo{author}{Srihari S~N}, \bibinfo{author}{Lam S~W}, \bibinfo{author}{Govindaraju V}, et~al.
\newblock \bibinfo{title}{Document image understanding.}
\newblock In: \bibinfo{booktitle}{FJCC}, \bibinfo{year}{1986}.
\newblock \bibinfo{pages}{87--95}

\bibitem{hwang2019post}
\bibinfo{author}{Hwang W}, \bibinfo{author}{Kim S}, \bibinfo{author}{Seo M}, et~al.
\newblock \bibinfo{title}{Post-{OCR} parsing: building simple and robust parser via bio tagging}.
\newblock In: \bibinfo{booktitle}{NeurIPS}, \bibinfo{year}{2019}

\bibitem{kim2022ocr}
\bibinfo{author}{Kim G}, \bibinfo{author}{Hong T}, \bibinfo{author}{Yim M}, et~al.
\newblock \bibinfo{title}{{OCR}-free document understanding transformer}.
\newblock In: \bibinfo{booktitle}{ECCV}, \bibinfo{year}{2022}.
\newblock \bibinfo{pages}{498--517}

\bibitem{luo2023geolayoutlm}
\bibinfo{author}{Luo C}, \bibinfo{author}{Cheng C}, \bibinfo{author}{Zheng Q}, et~al.
\newblock \bibinfo{title}{{GeoLayoutLM}: {Geometric} pre-training for visual information extraction}.
\newblock In: \bibinfo{booktitle}{CVPR}, \bibinfo{year}{2023}.
\newblock \bibinfo{pages}{7092--7101}

\bibitem{ye2023mplug}
\bibinfo{author}{Ye Q}, \bibinfo{author}{Xu H}, \bibinfo{author}{Xu G}, et~al.
\newblock \bibinfo{title}{{mPLUG-Owl}: {Modularization} empowers large language models with multimodality}.
\newblock \bibinfo{journal}{arXiv preprint arXiv:2304.14178}, \bibinfo{year}{2023}

\bibitem{feng2023unidoc}
\bibinfo{author}{Feng H}, \bibinfo{author}{Wang Z}, \bibinfo{author}{Tang J}, et~al.
\newblock \bibinfo{title}{{UniDoc}: A universal large multimodal model for simultaneous text detection, recognition, spotting and understanding}.
\newblock \bibinfo{journal}{arXiv preprint arXiv:2308.11592}, \bibinfo{year}{2023}

\bibitem{ye2023ureader}
\bibinfo{author}{Ye J}, \bibinfo{author}{Hu A}, \bibinfo{author}{Xu H}, et~al.
\newblock \bibinfo{title}{{UReader}: {Universal} {OCR}-free visually-situated language understanding with multimodal large language model}.
\newblock \bibinfo{journal}{arXiv preprint arXiv:2310.05126}, \bibinfo{year}{2023}

\bibitem{lv2023kosmos}
\bibinfo{author}{Lv T}, \bibinfo{author}{Huang Y}, \bibinfo{author}{Chen J}, et~al.
\newblock \bibinfo{title}{{KOSMOS}-2.5: A multimodal literate model}.
\newblock \bibinfo{journal}{arXiv preprint arXiv:2309.11419}, \bibinfo{year}{2023}

\bibitem{xu2021layoutxlm}
\bibinfo{author}{Xu Y}, \bibinfo{author}{Lv T}, \bibinfo{author}{Cui L}, et~al.
\newblock \bibinfo{title}{{LayoutxLM}: {Multimodal} pre-training for multilingual visually-rich document understanding}.
\newblock \bibinfo{journal}{arXiv preprint arXiv:2104.08836}, \bibinfo{year}{2021}

\bibitem{xu2020layoutlm}
\bibinfo{author}{Xu Y}, \bibinfo{author}{Li M}, \bibinfo{author}{Cui L}, et~al.
\newblock \bibinfo{title}{{LayoutLM}: {Pre-training} of text and layout for document image understanding}.
\newblock In: \bibinfo{booktitle}{KDD}, \bibinfo{year}{2020}.
\newblock \bibinfo{pages}{1192--1200}

\bibitem{huang2022layoutlmv3}
\bibinfo{author}{Huang Y}, \bibinfo{author}{Lv T}, \bibinfo{author}{Cui L}, et~al.
\newblock \bibinfo{title}{{LayoutLMv3}: {Pre-training} for document ai with unified text and image masking}.
\newblock In: \bibinfo{booktitle}{ACM MM}, \bibinfo{year}{2022}.
\newblock \bibinfo{pages}{4083--4091}

\bibitem{hong2022bros}
\bibinfo{author}{Hong T}, \bibinfo{author}{Kim D}, \bibinfo{author}{Ji M}, et~al.
\newblock \bibinfo{title}{Bros: A pre-trained language model focusing on text and layout for better key information extraction from documents}.
\newblock In: \bibinfo{booktitle}{AAAI}, \bibinfo{year}{2022}, volume~\bibinfo{volume}{36}.
\newblock \bibinfo{pages}{10767--10775}

\bibitem{bai2022wukong}
\bibinfo{author}{Bai H}, \bibinfo{author}{Liu Z}, \bibinfo{author}{Meng X}, et~al.
\newblock \bibinfo{title}{{Wukong-Reader}: Multi-modal pre-training for fine-grained visual document understanding}.
\newblock \bibinfo{journal}{arXiv preprint arXiv:2212.09621}, \bibinfo{year}{2022}

\bibitem{tang2023unifying}
\bibinfo{author}{Tang Z}, \bibinfo{author}{Yang Z}, \bibinfo{author}{Wang G}, et~al.
\newblock \bibinfo{title}{Unifying vision, text, and layout for universal document processing}.
\newblock In: \bibinfo{booktitle}{CVPR}, \bibinfo{year}{2023}.
\newblock \bibinfo{pages}{19254--19264}

\bibitem{li2021structext}
\bibinfo{author}{Li Y}, \bibinfo{author}{Qian Y}, \bibinfo{author}{Yu Y}, et~al.
\newblock \bibinfo{title}{{StrucTexT}: {Structured} text understanding with multi-modal transformers}.
\newblock In: \bibinfo{booktitle}{ACM MM}, \bibinfo{year}{2021}.
\newblock \bibinfo{pages}{1912--1920}

\bibitem{peng2022ernie}
\bibinfo{author}{Peng Q}, \bibinfo{author}{Pan Y}, \bibinfo{author}{Wang W}, et~al.
\newblock \bibinfo{title}{Ernie-layout: {Layout} knowledge enhanced pre-training for visually-rich document understanding}.
\newblock \bibinfo{journal}{arXiv preprint arXiv:2210.06155}, \bibinfo{year}{2022}

\bibitem{appalaraju2021docformer}
\bibinfo{author}{Appalaraju S}, \bibinfo{author}{Jasani B}, \bibinfo{author}{Kota B~U}, et~al.
\newblock \bibinfo{title}{{DocFormer}: End-to-end transformer for document understanding}.
\newblock In: \bibinfo{booktitle}{ICCV}, \bibinfo{year}{2021}.
\newblock \bibinfo{pages}{993--1003}

\bibitem{liao2020real}
\bibinfo{author}{Liao M}, \bibinfo{author}{Wan Z}, \bibinfo{author}{Yao C}, et~al.
\newblock \bibinfo{title}{Real-time scene text detection with differentiable binarization}.
\newblock In: \bibinfo{booktitle}{AAAI}, \bibinfo{year}{2020}, volume~\bibinfo{volume}{34}.
\newblock \bibinfo{pages}{11474--11481}

\bibitem{shi2016end}
\bibinfo{author}{Shi B}, \bibinfo{author}{Bai X}, \bibinfo{author}{Yao C}.
\newblock \bibinfo{title}{An end-to-end trainable neural network for image-based sequence recognition and its application to scene text recognition}.
\newblock \bibinfo{journal}{TPAMI}, \bibinfo{year}{2016}, \bibinfo{volume}{39}: \bibinfo{pages}{2298--2304}

\bibitem{zhang2023llavar}
\bibinfo{author}{Zhang Y}, \bibinfo{author}{Zhang R}, \bibinfo{author}{Gu J}, et~al.
\newblock \bibinfo{title}{{LLaVAR}: Enhanced visual instruction tuning for text-rich image understanding}.
\newblock \bibinfo{journal}{arXiv preprint arXiv:2306.17107}, \bibinfo{year}{2023}

\bibitem{ye2023mplug-doc}
\bibinfo{author}{Ye J}, \bibinfo{author}{Hu A}, \bibinfo{author}{Xu H}, et~al.
\newblock \bibinfo{title}{{mPLUG-DocOwl}: {Modularized} multimodal large language model for document understanding}.
\newblock \bibinfo{journal}{arXiv preprint arXiv:2307.02499}, \bibinfo{year}{2023}

\bibitem{lee2023pix2struct}
\bibinfo{author}{Lee K}, \bibinfo{author}{Joshi M}, \bibinfo{author}{Turc I~R}, et~al.
\newblock \bibinfo{title}{{Pix2Struct}: {Screenshot} parsing as pretraining for visual language understanding}.
\newblock In: \bibinfo{booktitle}{ICML}, \bibinfo{year}{2023}.
\newblock \bibinfo{pages}{18893--18912}

\bibitem{radford2021learning}
\bibinfo{author}{Radford A}, \bibinfo{author}{Kim J~W}, \bibinfo{author}{Hallacy C}, et~al.
\newblock \bibinfo{title}{Learning transferable visual models from natural language supervision}.
\newblock In: \bibinfo{booktitle}{ICML}, \bibinfo{year}{2021}.
\newblock \bibinfo{pages}{8748--8763}

\bibitem{liu2023hidden}
\bibinfo{author}{Liu Y}, \bibinfo{author}{Li Z}, \bibinfo{author}{Li H}, et~al.
\newblock \bibinfo{title}{On the hidden mystery of {OCR} in large multimodal models}.
\newblock \bibinfo{journal}{arXiv preprint arXiv:2305.07895}, \bibinfo{year}{2023}

\bibitem{vaswani2017attention}
\bibinfo{author}{Vaswani A}, \bibinfo{author}{Shazeer N}, \bibinfo{author}{Parmar N}, et~al.
\newblock \bibinfo{title}{Attention is all you need}.
\newblock \bibinfo{journal}{NeurIPS}, \bibinfo{year}{2017}, \bibinfo{volume}{30}

\bibitem{touvron2023llama}
\bibinfo{author}{Touvron H}, \bibinfo{author}{Lavril T}, \bibinfo{author}{Izacard G}, et~al.
\newblock \bibinfo{title}{{LLaMA}: {Open} and efficient foundation language models}.
\newblock \bibinfo{journal}{arXiv preprint arXiv:2302.13971}, \bibinfo{year}{2023}

\bibitem{chiang2023vicuna}
\bibinfo{author}{Chiang W~L}, \bibinfo{author}{Li Z}, \bibinfo{author}{Lin Z}, et~al.
\newblock \bibinfo{title}{Vicuna: An open-source chatbot impressing {GPT-4} with 90\%* {ChatGPT} quality}.
\newblock \bibinfo{journal}{See https://vicuna. lmsys. org}, \bibinfo{year}{2023}

\bibitem{ahmed1974discrete}
\bibinfo{author}{Ahmed N}, \bibinfo{author}{Natarajan T}, \bibinfo{author}{Rao K~R}.
\newblock \bibinfo{title}{Discrete cosine transform}.
\newblock \bibinfo{journal}{IEEE Transactions on Computers}, \bibinfo{year}{1974}, \bibinfo{volume}{100}: \bibinfo{pages}{90--93}

\bibitem{wallace1991jpeg}
\bibinfo{author}{Wallace G~K}.
\newblock \bibinfo{title}{The jpeg still picture compression standard}.
\newblock \bibinfo{journal}{Commun. ACM}, \bibinfo{year}{1991}, \bibinfo{volume}{34}: \bibinfo{pages}{30--44}

\bibitem{liu2023devil}
\bibinfo{author}{Liu H}, \bibinfo{author}{Jiang X}, \bibinfo{author}{Li X}, et~al.
\newblock \bibinfo{title}{The devil is in the frequency: Geminated gestalt autoencoder for self-supervised visual pre-training}.
\newblock In: \bibinfo{booktitle}{AAAI}, \bibinfo{year}{2023}, volume~\bibinfo{volume}{37}.
\newblock \bibinfo{pages}{1649--1656}

\bibitem{liu2022nommer}
\bibinfo{author}{Liu H}, \bibinfo{author}{Jiang X}, \bibinfo{author}{Li X}, et~al.
\newblock \bibinfo{title}{{NomMer}: Nominate synergistic context in vision transformer for visual recognition}.
\newblock In: \bibinfo{booktitle}{CVPR}, \bibinfo{year}{2022}.
\newblock \bibinfo{pages}{12073--12082}

\bibitem{liu2018fots}
\bibinfo{author}{Liu X}, \bibinfo{author}{Liang D}, \bibinfo{author}{Yan S}, et~al.
\newblock \bibinfo{title}{{FOTS}: Fast oriented text spotting with a unified network}.
\newblock In: \bibinfo{booktitle}{CVPR}, \bibinfo{year}{2018}.
\newblock \bibinfo{pages}{5676--5685}

\bibitem{hossain2019comprehensive}
\bibinfo{author}{Hossain M~Z}, \bibinfo{author}{Sohel F}, \bibinfo{author}{Shiratuddin M~F}, et~al.
\newblock \bibinfo{title}{A comprehensive survey of deep learning for image captioning}.
\newblock \bibinfo{journal}{ACM Computing Surveys}, \bibinfo{year}{2019}, \bibinfo{volume}{51}: \bibinfo{pages}{1--36}

\bibitem{brown2020language}
\bibinfo{author}{Brown T}, \bibinfo{author}{Mann B}, \bibinfo{author}{Ryder N}, et~al.
\newblock \bibinfo{title}{Language models are few-shot learners}.
\newblock \bibinfo{journal}{NeurIPS}, \bibinfo{year}{2020}, \bibinfo{volume}{33}: \bibinfo{pages}{1877--1901}

\bibitem{liu2023visual}
\bibinfo{author}{Liu H}, \bibinfo{author}{Li C}, \bibinfo{author}{Wu Q}, et~al.
\newblock \bibinfo{title}{Visual instruction tuning}.
\newblock \bibinfo{journal}{arXiv preprint arXiv:2304.08485}, \bibinfo{year}{2023}

\bibitem{yu2023structextv2}
\bibinfo{author}{Yu Y}, \bibinfo{author}{Li Y}, \bibinfo{author}{Zhang C}, et~al.
\newblock \bibinfo{title}{{StrucTexTv2}: {Masked} visual-textual prediction for document image pre-training}.
\newblock \bibinfo{journal}{arXiv preprint arXiv:2303.00289}, \bibinfo{year}{2023}

\bibitem{he2022masked}
\bibinfo{author}{He K}, \bibinfo{author}{Chen X}, \bibinfo{author}{Xie S}, et~al.
\newblock \bibinfo{title}{Masked autoencoders are scalable vision learners}.
\newblock In: \bibinfo{booktitle}{CVPR}, \bibinfo{year}{2022}.
\newblock \bibinfo{pages}{16000--16009}

\bibitem{devlin2018bert}
\bibinfo{author}{Devlin J}, \bibinfo{author}{Chang M~W}, \bibinfo{author}{Lee K}, et~al.
\newblock \bibinfo{title}{{BERT}: Pre-training of deep bidirectional transformers for language understanding}.
\newblock \bibinfo{journal}{arXiv preprint arXiv:1810.04805}, \bibinfo{year}{2018}

\bibitem{peng2023kosmos}
\bibinfo{author}{Peng Z}, \bibinfo{author}{Wang W}, \bibinfo{author}{Dong L}, et~al.
\newblock \bibinfo{title}{{KOSMOS-2}: Grounding multimodal large language models to the world}.
\newblock \bibinfo{journal}{arXiv preprint arXiv:2306.14824}, \bibinfo{year}{2023}

\bibitem{li2023monkey}
\bibinfo{author}{Li Z}, \bibinfo{author}{Yang B}, \bibinfo{author}{Liu Q}, et~al.
\newblock \bibinfo{title}{Monkey: Image resolution and text label are important things for large multi-modal models}.
\newblock In: \bibinfo{booktitle}{CVPR}, \bibinfo{year}{2024}

\bibitem{liu2024textmonkey}
\bibinfo{author}{Liu Y}, \bibinfo{author}{Yang B}, \bibinfo{author}{Liu Q}, et~al.
\newblock \bibinfo{title}{{TextMonkey}: An {OCR}-free large multimodal model for understanding document}.
\newblock \bibinfo{journal}{arXiv preprint arXiv:2403.04473}, \bibinfo{year}{2024}

\bibitem{liu2021swin}
\bibinfo{author}{Liu Z}, \bibinfo{author}{Lin Y}, \bibinfo{author}{Cao Y}, et~al.
\newblock \bibinfo{title}{Swin transformer: Hierarchical vision transformer using shifted windows}.
\newblock In: \bibinfo{booktitle}{ICCV}, \bibinfo{year}{2021}.
\newblock \bibinfo{pages}{10012--10022}

\bibitem{zhu2023minigpt}
\bibinfo{author}{Zhu D}, \bibinfo{author}{Chen J}, \bibinfo{author}{Shen X}, et~al.
\newblock \bibinfo{title}{{MiniGPT-4}: Enhancing vision-language understanding with advanced large language models}.
\newblock \bibinfo{journal}{arXiv preprint arXiv:2304.10592}, \bibinfo{year}{2023}

\bibitem{wang2011end}
\bibinfo{author}{Wang K}, \bibinfo{author}{Babenko B}, \bibinfo{author}{Belongie S}.
\newblock \bibinfo{title}{End-to-end scene text recognition}.
\newblock In: \bibinfo{booktitle}{ICCV}, \bibinfo{year}{2011}.
\newblock \bibinfo{pages}{1457--1464}

\bibitem{li2023blip}
\bibinfo{author}{Li J}, \bibinfo{author}{Li D}, \bibinfo{author}{Savarese S}, et~al.
\newblock \bibinfo{title}{Blip-2: Bootstrapping language-image pre-training with frozen image encoders and large language models}.
\newblock In: \bibinfo{booktitle}{ICML}, \bibinfo{year}{2023}.
\newblock \bibinfo{pages}{19730--19742}

\bibitem{dai2024instructblip}
\bibinfo{author}{Dai W}, \bibinfo{author}{Li J}, \bibinfo{author}{Li D}, et~al.
\newblock \bibinfo{title}{Instructblip: Towards general-purpose vision-language models with instruction tuning}.
\newblock \bibinfo{journal}{NeurIPS}, \bibinfo{year}{2024}, \bibinfo{volume}{36}

\bibitem{hu2024bliva}
\bibinfo{author}{Hu W}, \bibinfo{author}{Xu Y}, \bibinfo{author}{Li Y}, et~al.
\newblock \bibinfo{title}{Bliva: A simple multimodal {LLM} for better handling of text-rich visual questions}.
\newblock In: \bibinfo{booktitle}{AAAI}, \bibinfo{year}{2024}, volume~\bibinfo{volume}{38}.
\newblock \bibinfo{pages}{2256--2264}

\bibitem{liu2023improved}
\bibinfo{author}{Liu H}, \bibinfo{author}{Li C}, \bibinfo{author}{Li Y}, et~al.
\newblock \bibinfo{title}{Improved baselines with visual instruction tuning}.
\newblock \bibinfo{journal}{arXiv preprint arXiv:2310.03744}, \bibinfo{year}{2023}

\bibitem{wang2023towards}
\bibinfo{author}{Wang Y}, \bibinfo{author}{Zhou W}, \bibinfo{author}{Feng H}, et~al.
\newblock \bibinfo{title}{Towards improving document understanding: An exploration on text-grounding via {MLLMs}}.
\newblock \bibinfo{journal}{arXiv preprint arXiv:2311.13194}, \bibinfo{year}{2023}

\bibitem{ye2023mplug2}
\bibinfo{author}{Ye Q}, \bibinfo{author}{Xu H}, \bibinfo{author}{Ye J}, et~al.
\newblock \bibinfo{title}{{mPLUG-owl2}: Revolutionizing multi-modal large language model with modality collaboration}.
\newblock In: \bibinfo{booktitle}{CVPR}, \bibinfo{year}{2024}

\bibitem{dong2024internlm}
\bibinfo{author}{Dong X}, \bibinfo{author}{Zhang P}, \bibinfo{author}{Zang Y}, et~al.
\newblock \bibinfo{title}{{InternLM-XComposer2}: Mastering free-form text-image composition and comprehension in vision-language large model}.
\newblock \bibinfo{journal}{arXiv preprint arXiv:2401.16420}, \bibinfo{year}{2024}

\bibitem{mathew2021docvqa}
\bibinfo{author}{Mathew M}, \bibinfo{author}{Karatzas D}, \bibinfo{author}{Jawahar C}.
\newblock \bibinfo{title}{{DocVQA}: A dataset for {VQA} on document images}.
\newblock In: \bibinfo{booktitle}{WACV}, \bibinfo{year}{2021}.
\newblock \bibinfo{pages}{2200--2209}

\bibitem{mishra2019ocr}
\bibinfo{author}{Mishra A}, \bibinfo{author}{Shekhar S}, \bibinfo{author}{Singh A~K}, et~al.
\newblock \bibinfo{title}{{OCR-VQA}: Visual question answering by reading text in images}.
\newblock In: \bibinfo{booktitle}{ICDAR}, \bibinfo{year}{2019}.
\newblock \bibinfo{pages}{947--952}

\bibitem{singh2019towards}
\bibinfo{author}{Singh A}, \bibinfo{author}{Natarajan V}, \bibinfo{author}{Shah M}, et~al.
\newblock \bibinfo{title}{Towards {VQA} models that can read}.
\newblock In: \bibinfo{booktitle}{CVPR}, \bibinfo{year}{2019}.
\newblock \bibinfo{pages}{8317--8326}

\bibitem{mathew2022infographicvqa}
\bibinfo{author}{Mathew M}, \bibinfo{author}{Bagal V}, \bibinfo{author}{Tito R}, et~al.
\newblock \bibinfo{title}{Infographic{VQA}}.
\newblock In: \bibinfo{booktitle}{WACV}, \bibinfo{year}{2022}.
\newblock \bibinfo{pages}{1697--1706}

\bibitem{masry2022chartqa}
\bibinfo{author}{Masry A}, \bibinfo{author}{Long D~X}, \bibinfo{author}{Tan J~Q}, et~al.
\newblock \bibinfo{title}{{ChartQA}: A benchmark for question answering about charts with visual and logical reasoning}.
\newblock \bibinfo{journal}{arXiv preprint arXiv:2203.10244}, \bibinfo{year}{2022}

\bibitem{kahou2017figureqa}
\bibinfo{author}{Kahou S~E}, \bibinfo{author}{Michalski V}, \bibinfo{author}{Atkinson A}, et~al.
\newblock \bibinfo{title}{{FigureQA}: An annotated figure dataset for visual reasoning}.
\newblock \bibinfo{journal}{arXiv preprint arXiv:1710.07300}, \bibinfo{year}{2017}

\bibitem{jaume2019funsd}
\bibinfo{author}{Jaume G}, \bibinfo{author}{Ekenel H~K}, \bibinfo{author}{Thiran J~P}.
\newblock \bibinfo{title}{Funsd: A dataset for form understanding in noisy scanned documents}.
\newblock In: \bibinfo{booktitle}{ICDARW}, \bibinfo{year}{2019}, volume~\bibinfo{volume}{2}.
\newblock \bibinfo{pages}{1--6}

\bibitem{huang2019icdar2019}
\bibinfo{author}{Huang Z}, \bibinfo{author}{Chen K}, \bibinfo{author}{He J}, et~al.
\newblock \bibinfo{title}{{ICDAR} 2019 competition on scanned receipt {OCR} and information extraction}.
\newblock In: \bibinfo{booktitle}{ICDAR}, \bibinfo{year}{2019}.
\newblock \bibinfo{pages}{1516--1520}

\bibitem{kuang2023visual}
\bibinfo{author}{Kuang J}, \bibinfo{author}{Hua W}, \bibinfo{author}{Liang D}, et~al.
\newblock \bibinfo{title}{Visual information extraction in the wild: practical dataset and end-to-end solution}.
\newblock In: \bibinfo{booktitle}{ICDAR}, \bibinfo{year}{2023}.
\newblock \bibinfo{pages}{36--53}

\bibitem{smith2019super}
\bibinfo{author}{Smith L~N}, \bibinfo{author}{Topin N}.
\newblock \bibinfo{title}{Super-convergence: Very fast training of neural networks using large learning rates}.
\newblock In: \bibinfo{booktitle}{AIMLMOA}, \bibinfo{year}{2019}, volume \bibinfo{volume}{11006}.
\newblock \bibinfo{pages}{369--386}

\bibitem{loshchilov2017decoupled}
\bibinfo{author}{Loshchilov I}, \bibinfo{author}{Hutter F}.
\newblock \bibinfo{title}{Decoupled weight decay regularization}.
\newblock \bibinfo{journal}{arXiv preprint arXiv:1711.05101}, \bibinfo{year}{2017}

\bibitem{biten2019icdar}
\bibinfo{author}{Biten A~F}, \bibinfo{author}{Tito R}, \bibinfo{author}{Mafla A}, et~al.
\newblock \bibinfo{title}{{ICDAR} 2019 competition on scene text visual question answering}.
\newblock In: \bibinfo{booktitle}{ICDAR}, \bibinfo{year}{2019}.
\newblock \bibinfo{pages}{1563--1570}

\bibitem{Qwen-VL}
\bibinfo{author}{Bai J}, \bibinfo{author}{Bai S}, \bibinfo{author}{Yang S}, et~al.
\newblock \bibinfo{title}{Qwen-{VL}: A frontier large vision-language model with versatile abilities}.
\newblock \bibinfo{journal}{arXiv preprint arXiv:2308.12966}, \bibinfo{year}{2023}

\bibitem{dosovitskiy2020image}
\bibinfo{author}{Dosovitskiy A}, \bibinfo{author}{Beyer L}, \bibinfo{author}{Kolesnikov A}, et~al.
\newblock \bibinfo{title}{An image is worth 16x16 words: Transformers for image recognition at scale}.
\newblock \bibinfo{journal}{arXiv preprint arXiv:2010.11929}, \bibinfo{year}{2020}

\bibitem{tito2023hierarchical}
\bibinfo{author}{Tito R}, \bibinfo{author}{Karatzas D}, \bibinfo{author}{Valveny E}.
\newblock \bibinfo{title}{Hierarchical multimodal transformers for multipage {DocVQA}}.
\newblock \bibinfo{journal}{PR}, \bibinfo{year}{2023}, \bibinfo{volume}{144}: \bibinfo{pages}{109834}

\end{thebibliography}

\end{document}